%% file: main.tex
\documentclass{article}
\usepackage{spconf,amsmath,graphicx}
\usepackage[symbol]{footmisc}

\title{Fast and Accurate Scene Parsing via Bi-direction Alignment Networks}
\name{Yanran Wu$^{1}$ $^+$, Xiangtai Li$^{2}$ $^+$, Chen Shi$^{1}$, Yunhai Tong$^{2}$, Yang Hua$^{3}$, Tao Song$^{1\dagger}$\thanks{$^{\dagger}$Corresponding Author, E-mail: songt333@sjtu.edu.cn. $^+$ The first two authors contribute equally. This work is partially funded by National Natural Science Foundation of China (NO. 61872234, 61732010, 61525204), Shanghai Key Laboratory of Scalable Computing and Systems.}, Ruhui Ma$^{1}$, Haibing Guan$^{1}$}
\address{$^{1}$Shanghai Jiao Tong University, China \\
        $^{2}$ Peking University, China \\
        $^{3}$Queen’s University Belfast, UK
        }
\usepackage{amsfonts,amssymb}
\usepackage{bbm}
\usepackage{array,multirow}
\usepackage{threeparttable}
\usepackage{adjustbox}
\usepackage{colortbl}
\usepackage{graphicx}
\usepackage{booktabs}
\usepackage{subfigure} 
\usepackage{url}
\usepackage[dvipsnames]{xcolor}

\makeatletter
\newcommand\thefontsize{The current font size is: \f@size pt}
\makeatother
\begin{document}

\maketitle

\begin{abstract}
In this paper, we propose an effective method for fast and accurate scene parsing called Bidirectional Alignment Network (BiAlignNet). Previously, one representative work BiSeNet~\cite{bisenet} uses two different paths (Context Path and Spatial Path) to achieve balanced learning of semantics and details, respectively. However, the relationship between the two paths is not well explored. We argue that both paths can benefit each other in a complementary way. Motivated by this, we propose a novel network by aligning two-path information into each other through a learned flow field. To avoid the noise and semantic gaps, we introduce a Gated Flow Alignment Module to align both features in a bidirectional way. Moreover, to make the Spatial Path learn more detailed information, we present an edge-guided hard pixel mining loss to supervise the aligned learning process. Our method achieves 80.1\% and 78.5\% mIoU in validation and test set of Cityscapes while running at 30 FPS with full resolution inputs. Code and models will be available at \url{https://github.com/jojacola/BiAlignNet}.

\end{abstract}

\begin{keywords}
Bidirectional Alignment Network, Fast and Accurate Scene Parsing
\end{keywords}

\input{1introduction}
\input{3method}
\input{4experiment}

\input{5conclusion}

\small{
\bibliographystyle{IEEEbib}
\bibliography{egbib}
}

\end{document}

%% file: 1introduction.tex
\section{Introduction}
\label{sec:intro}





\begin{figure}[t]
	\centering  
	\vspace{-0.35cm} 
	\subfigtopskip=2pt 
	\subfigbottomskip=2pt 
	\subfigcapskip=-5pt 
	\subfigure[Atrous Conv\cite{dilation}]{
		\label{fig:sub_astous}
		\includegraphics[width=0.4\linewidth]{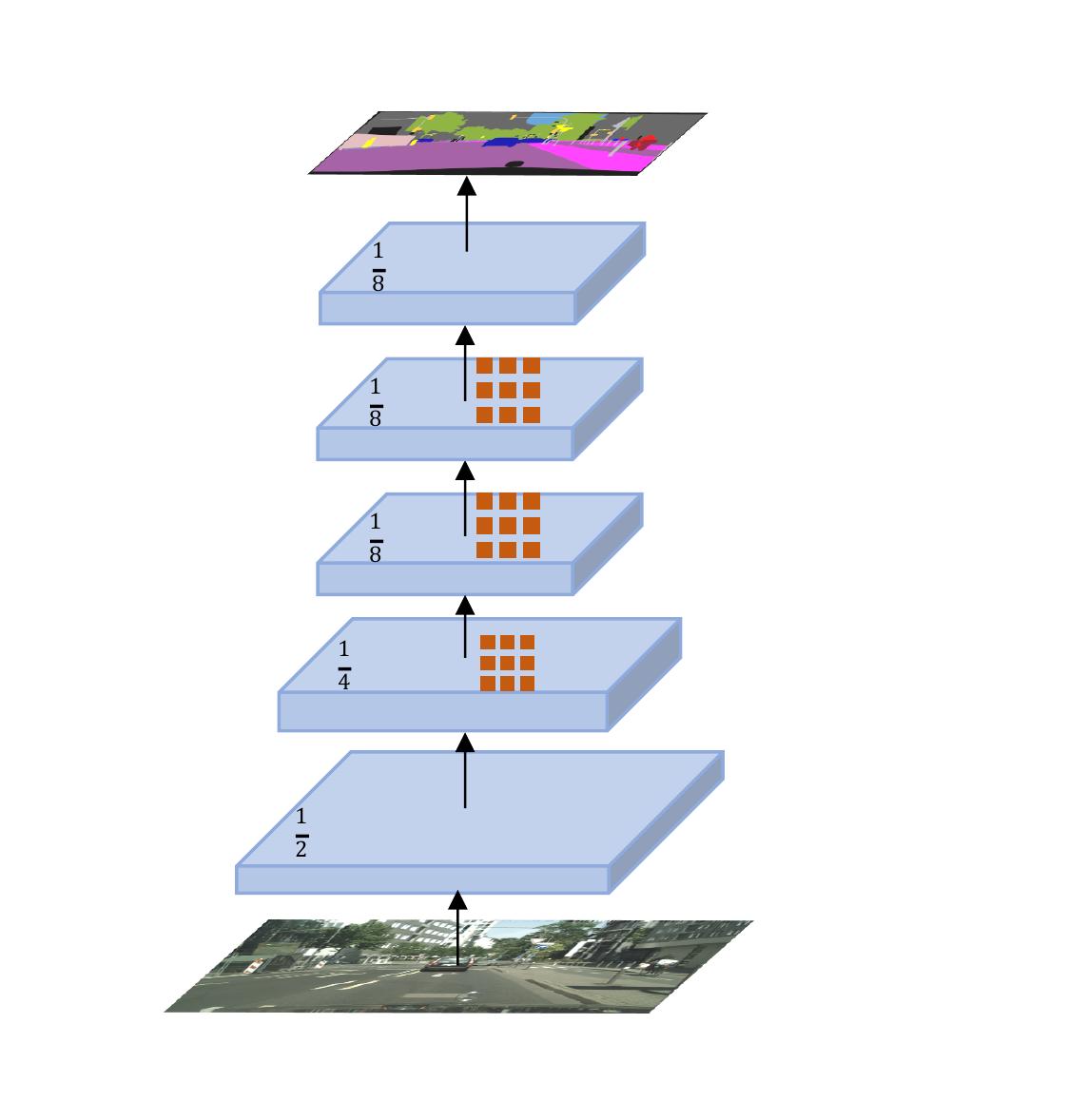}}
	\subfigure[FPN\cite{fpn}]{
		\label{fig:sub_fpn}
		\includegraphics[width=0.4\linewidth]{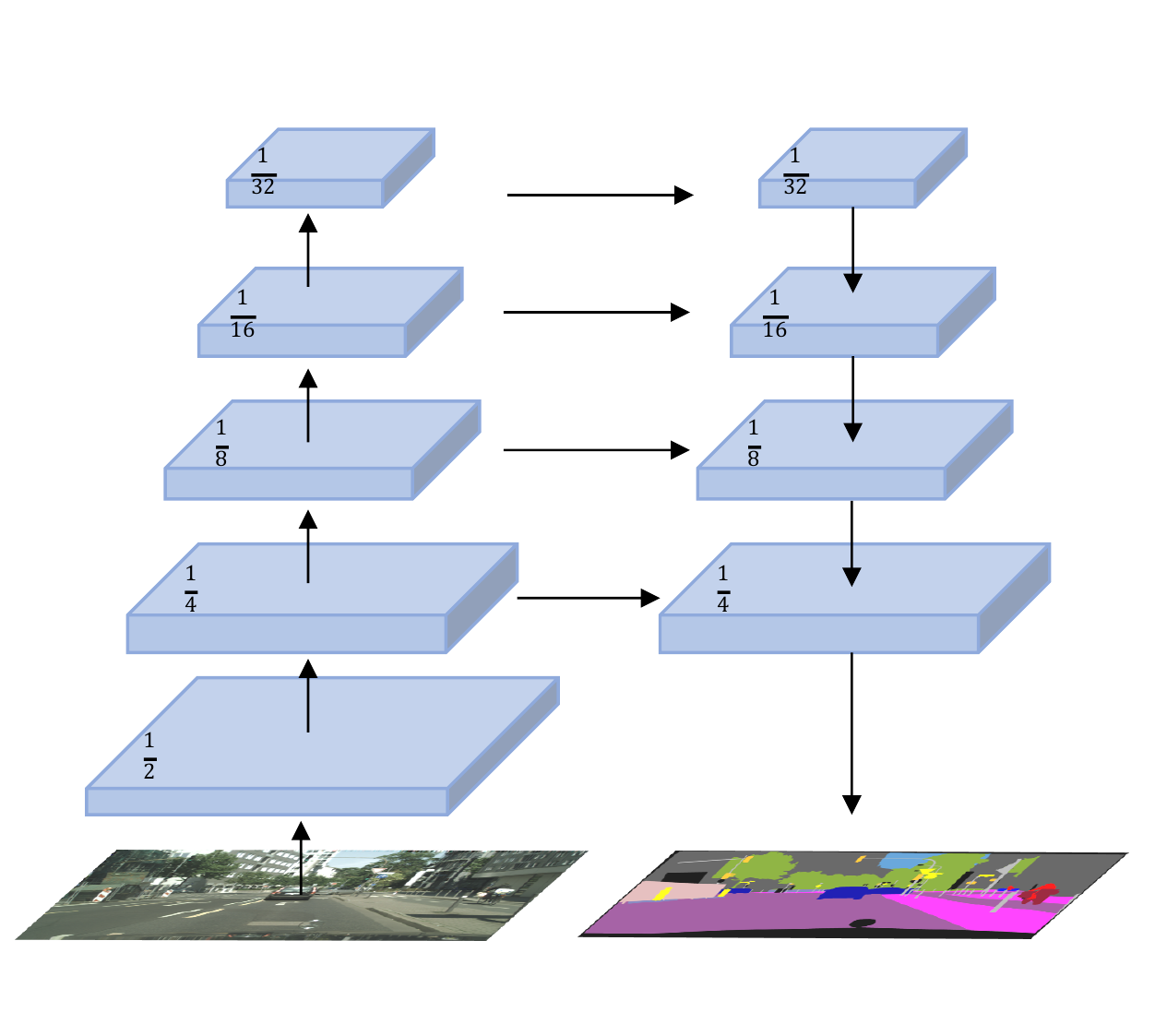}}
	\subfigure[BiSeNet\cite{bisenet}]{
		\label{fig:sub_bisenet}
		\includegraphics[width=0.35\linewidth]{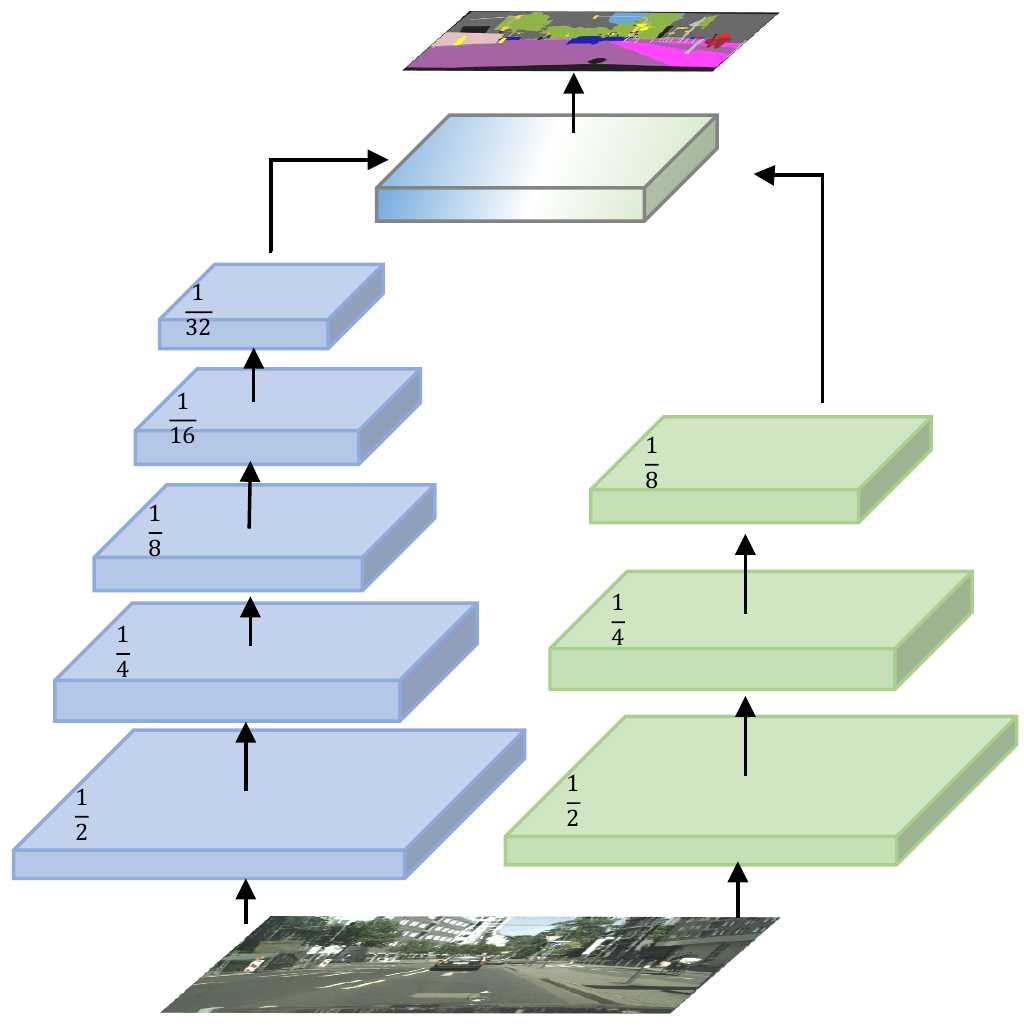}}
	\subfigure[Proposed BiAlignNet]{
		\label{fig:sub_bialign}
		\includegraphics[width=0.35\linewidth]{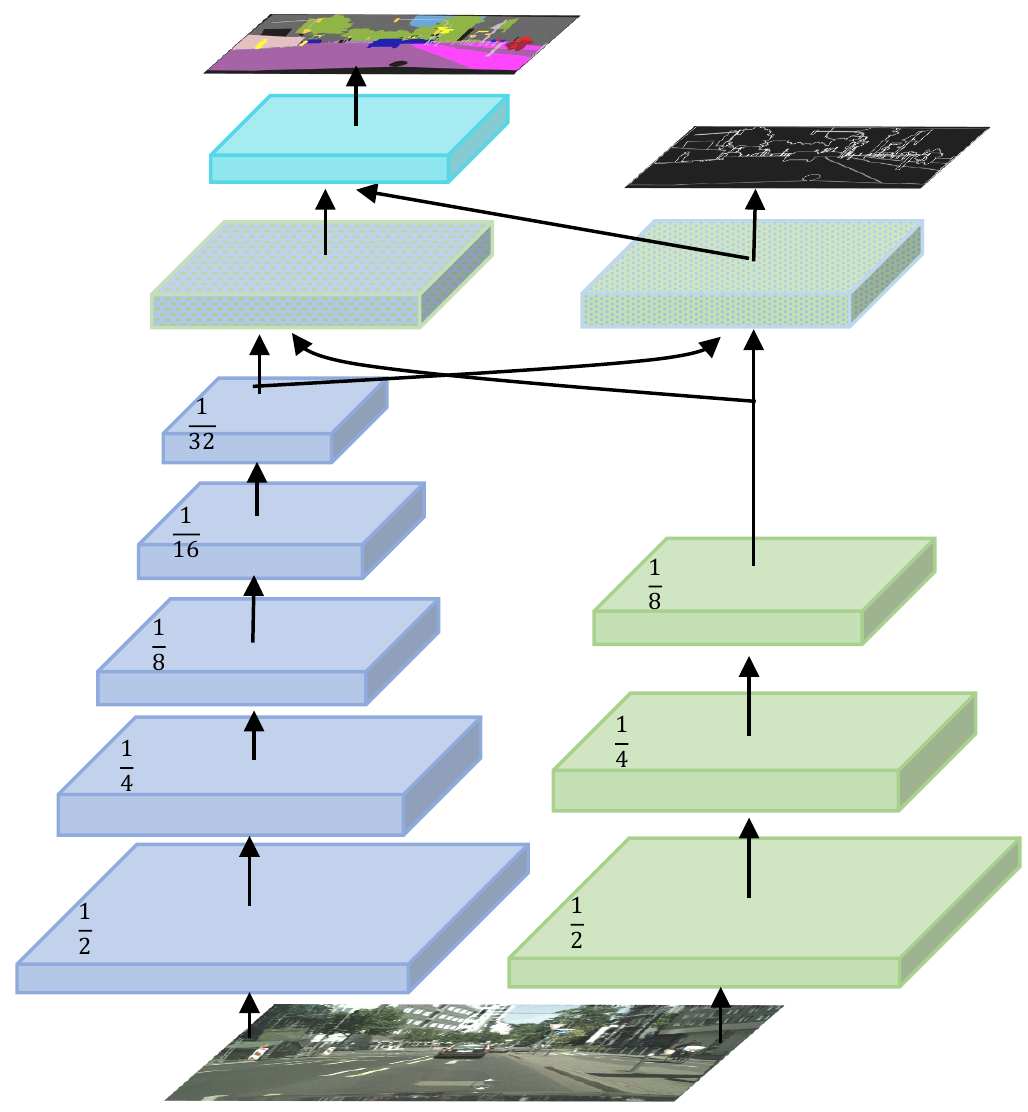}}
	\caption{\textbf{Comparison of different segmentation architectures.} \subref{fig:sub_astous} uses astrous convolution layers to obtain larger receptive field and high resolution feature map but introduces heavy computation complexity. \subref{fig:sub_fpn} is a FPN-like model. It gets a high resolution feature map by adding top-down and lateral fusions. \subref{fig:sub_bisenet} shows the structure of BiSeNet\cite{bisenet}. We propose \subref{fig:sub_bialign} to maximize the utilization between two paths and add different supervision according to their priorities. Best viewed in color. }
	\label{fig:teaser}
\end{figure}

Semantic Segmentation is a fundamental vision task that aims to classify each pixel in the images correctly. Some earlier approaches~\cite{deeplabv1, li2011superpixel} use structured prediction operators such as conditional random fields (CRFs) to refine segmentation results. Recent methods for semantic segmentation are predominantly based on FCNs~\cite{fcn}. Current state-of-the-art methods~\cite{pspnet,DAnet,nvidia_seg_video} apply atrous convolutions~\cite{dilation} at the last several stages of their networks to yield feature maps with strong semantic representation while at the same time maintaining the high resolution, as shown in Fig.~\ref{fig:teaser}(a). Moreover, there are also several methods based on Feature Pyramid Network (FPN)-like~\cite{fpn,PanopticFPN,unet} models which leverage the lateral path to fuse feature maps in a top-down manner. In this way, the deep features of the last several layers strengthen the shallow features with high resolution. Therefore, the refined features are possible to keep high resolution and meanwhile catch semantic representation, which is beneficial to the accuracy improvement, as shown in Fig.~\ref{fig:teaser}(b). However, both designs are not practical for real-time settings. The former methods~\cite{pspnet,DAnet} require extra computation since the feature maps in the last stages can reach up to 64 times bigger than those in FCNs. Meanwhile, the latter one~\cite{PanopticFPN} has a heavier fusion operation in their decoder. For example, under a single GTX 1080Ti GPU, the previous model PSPNet~\cite{pspnet} has a frame rate of only 1.6 FPS for $1024 \times 2048$ input images. As a consequence, this is very problematic for many time-critical applications, such as autonomous driving and robot navigation, which desperately demand real-time online data processing.

There are several specific designed real-time semantic segmentation models~\cite{ICnet,dfanet,bisenet,bisenetv2} handling above issues. However, these methods can not achieve satisfactory segmentation results as accurate models. The representative works BiSeNets~\cite{bisenet,bisenetv2} propose to use two different paths for learning spatial details and coarse context information respectively, shown in Fig.~\ref{fig:teaser}(c). 
However, they have not explored the interaction between two data flows explicitly. We believe such two data flows contain complementary content that can benefit each other. In this paper, we propose a new network architecture for real-time scene parsing settings. As shown in Fig.~\ref{fig:teaser}(d), two paths interact with each other through specific design modules before the fusing. Motivated by a recent alignment module~\cite{sfnet} which deforms the entire feature map using a learned flow field, we propose a Gated Flow Alignment Module to avoid noise during the fusing since two paths contain diverse information. The proposed module is light-weight and can be inserted on each path before fusion.
The features are aligned to each other through the learned flow fields.
Moreover, to make the spatial path learn detailed information, we supervise it using the edge-guided hard pixel mining loss~\cite{ohem} to further improve the performance. We term our network as BiAlignNet for short.

Finally, we evaluate BiAlignNet on two datasets, i.e.,  Cityscapes~\cite{Cityscapes} and CamVid~\cite{CamVid}. The results demonstrate the effectiveness of the proposed components. Specifically, our methods improve the origin BiSegNet baseline by about 2\% mIoU on the test set of Cityscapes with only 3 FPS drop. Our method can achieve 78.5\% mIoU while running at 32 FPS on single 1080Ti without acceleration.


%% file: 3method.tex
\section{Method}
\label{Method}
\begin{figure}[t]
    \centering
    \includegraphics[width=\linewidth]{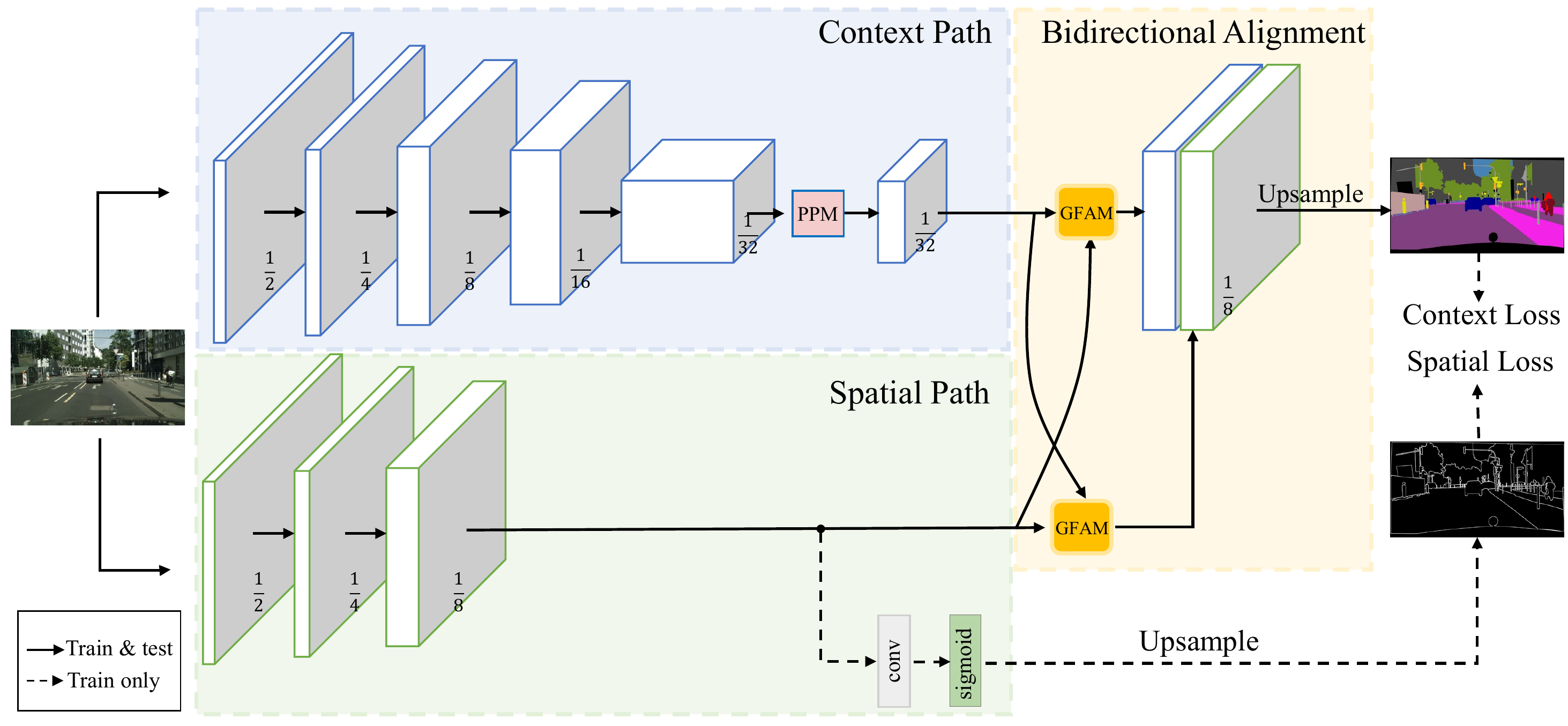}
    \caption{\textbf{Overview of the BiAlignNet.} The context path is in the \textcolor{blue}{blue} box. The spatial path is in the \textcolor{YellowGreen}{green} box.  \textcolor{Dandelion}{Orange} part represents the bidirectional alignment. Best viewed in color.}
    \label{fig:bialgin}
\end{figure}

We present the overall network architecture in Fig.~\ref{fig:bialgin}. BiAlignNet includes the following three parts: two pathways, which are Spatial Path and Context Path, and Bidirectional Alignment using Gated Flow Alignment Module to align features in both directions. We also specially design the loss functions explained in Sec.~\ref{sec:loss} to supervise different sorts of information in two paths at last.


\subsection{Spatial Path and Context Path}
We briefly review the spatial and context path in BiSeNet~\cite{bisenet}. The spatial path is designed to capture the low-level information from the input image. We only use shallow layers to preserve spatial details. It only consists of three convolution layers with batch normalization and ReLU. Each layer has a stride of 2, so the final feature map of the spatial path is $\frac{1}{8}$ of the input size. The context path is responsible for extracting high-level information using a deeper network with more downsample operation. For implementation, we employ lightweight backbone DFNet~\cite{DF-seg-net} series for context path. Pyramid Pooling Module (PPM)~\cite{pspnet}, which has shown a strong ability to catch contextual information, is also added to our model. All backbones have four stages of residual blocks, and the first layer of each stage has a stride of 2. Thus, the final output of the context path is $\frac{1}{32}$ of the input size. 

\subsection{Bidirectional Alignment}
\label{sec:gfam}
In this section, we present a Gated Flow Alignment Module (GFAM) to align features with each other. The original FAM~\cite{sfnet} is proposed to align adjacent features in the decoder. However, directly using such a module may lead to inferior results because of the huge semantic gap between the two paths. Thus, we plug a gate into the FAM to avoid the noises and highlight the important information.
%
Suppose $\mathbf{F}_s$ is the source feature, and we want to align the information from $\mathbf{F}_s$ to target feature $\mathbf{F}_t$. Inspired by original FAM~\cite{sfnet}, we first generate a flow field grid $G$:

\begin{equation}
    G = conv(cat(\mathbf{F}_s || \mathbf{F}_t)),
\end{equation}
where $\mathbf{F}_s$ and $\mathbf{F}_t$ can be features from the spatial path and the context path respectively, and vice versa. The feature map that has a smaller size is bilinearly upsampled to reach the same size as the larger one. 

After flow field grid generation, we adopt a pixel-wise gate to emphasize the important part in current data flow:
\begin{equation}
    \hat{G} = \sigma(conv(\mathbf{F}_t)) \odot G,
\end{equation}
where $\hat{G}$ is the gated flow field grid, $\sigma$ means the sigmoid layer and $\odot$ represents element-wise product. 

Each position $p$ in target feature $\mathbf{F}_t$ can be mapped to a position $p^\prime$, according to the values in gated flow field grid $\hat{G}$. Note that the mapping result is not an integer, so the value at $\mathbf{F}_t(p^\prime)$ is interpolated by the values of the 4-neighbors $\mathcal{N}\left(p^\prime\right)$ (top-left, top-right, bottom-left, and bottom-right):
\begin{equation}
    \hat{\mathbf{F}_t}\left(p\right)=\sum_{i \in \mathcal{N}\left(p^\prime\right)} w_{p} \mathbf{F}_t(p^\prime), 
\end{equation}
where $w_{p}$ is the bilinear kernel weights estimated by the distance of warped grid, $\hat{\mathbf{F}_t}$ is the target feature aligned with information from source feature $\mathbf{F}_s$. In BiAlignNet, we take both spatial feature and context feature as source features to align with each other bidirectionally. In this way, different pieces of information can complement each other, as shown in the orange box of Fig.~\ref{fig:bialgin}.

\subsection{Loss Function}
\label{sec:loss}
The spatial path gives priority to spatial details while context path focuses on high-level semantic context. 
To force spatial path to focus on detailed information, we introduce an edge-guided hard pixel indicator map $d$ to supervise the learning. $d$ is predicted from the spatial path feature and normalized by a sigmoid layer. Since most of the fine information are concentrated in the boundaries, the edge map $b$ is derived from the segmentation labels through algorithm~\cite{findCountour} which retrieves contours from the binary image. We utilize the edge map $b$ to guide the prediction of indicator $d$.
As for context path, we use cross-entropy loss with online hard example mining (OHEM)~\cite{ohem,bisenet}. We jointly supervise two paths with a loss function $L$:
\begin{equation}
    L = L_{spatial}(d, b, s, g) + L_{context}(s, g),
\end{equation}
where $s$ is the predicted segmentation output of the model and $g$ is the ground truth segmentation labels, and $L_{context}$ is the OHEM loss. $L_{spatial}$ is calculated from the following equation. 
\begin{equation}
    L_{spatial} =\lambda L_{bce}(d, b) + L_{hard}(s, g, d), 
\end{equation}
\begin{equation}
    L_{hard} = -\frac{1}{K} \sum_{i=1}^{N} \mathbbm{1}\left[s_{i, g_{i}}<t_{K} \& d_{i}>t_{b}\right] \cdot \log s_{i, g_{i}},
    \label{eq:hard}
\end{equation}
where $L_{bce}$ is the binary cross-entropy loss for edge-guided hard pixel indicator $d$, $L_{hard}$ mines the hard pixels with high probability in $d$ and calculate the cross-entropy loss. $N$ is the total number of pixels. $\mathbbm{1}[x]=1$ if $x=1$ otherwise 0. First Eq.~\ref{eq:hard} filters the positions that have a higher probability than threshold $t_b$=0.8 in $d$. Then it picks positions within top $K$ losses, where $t_K$ is the threshold for top $K$ loss. Empirically, we set $\lambda= 25$ to balance the losses in all experiments. In this way, the spatial path learns more detailed information during the training.

%% file: 4experiment.tex
\section{Experiment}
\label{exp}
\subsection{Datasets}
We carry out experiments on Cityscapes and Camvid datasets.
Cityscapes~\cite{Cityscapes} is a large street scene dataset which contains 2,975 fine-annotated images for training, 500 images for validation and a testing set without annotations of 1,525 images. All images in this dataset have a high resolution of 1,024$\times$2,048. CamVid~\cite{CamVid} is another road scene dataset. This dataset contains 367 training images, 101 validation images and 233 testing images with a resolution of $720 \times 960$. 

\subsection{Speed and Accuracy Analysis}
\textbf{Implementation Details.} Our experiments are done with the PyTorch framework. We use stochastic gradient descent (SGD) with a batch size of 16 and a momentum of 0.9 and weight decay of 5e-4. The initial learning rate is 0.01 with a "poly" learning rate strategy in which the initial rate is multiplied by $\left(1-\frac{\text{ iter }}{\text{total\_iter}}\right)^{0.9}$. As for data augmentation, we randomly horizontally flip the images and randomly resize them with a scale of [0.5, 2.0], and crop images to a size of 1024$\times$1024 (720$\times$720 for CamVid). We use the single scale inference and report the speed with one 1080Ti GPU.

\begin{table}[!t]\setlength{\tabcolsep}{6pt}
\caption{\textbf{Comparison on Cityscapes {\it val} and {\it test} set with state-of-the-art real-time models.}   Notation: $\gamma$ is the downsampling ratio corresponding to the original $1024\times 2048$  resolution, for example, $\gamma=0.75$ means the model's input size is $768 \times 1536$. "*" noted methods and ours are tested on single 1080Ti GPU.}
	\centering
		\label{table:cityscapes_sota_speed_acc2}
	\begin{threeparttable}
		\scalebox{0.70}{
\begin{tabular}{lcccccc}
\toprule[0.2em]
\multirow{2}{*}{Method} & \multirow{2}{*}{$\gamma$} & \multirow{2}{*}{Backbone} & \multicolumn{2}{c}{mIoU ($\%$) } & \multirow{2}{*}{\#FPS} & \multirow{2}{*}{\#Params} \\ \cline{4-5}
                        &                        &                           & val         & test        &                      &                         \\ \toprule[0.2em]
ENet~\cite{ENnet}         & 0.5 &    -        & -   &58.3    &    60  &   0.4M\\ 
ESPNet~\cite{ESPNet}      & 0.5 & ESPNet      & -   &60.3    &  132   &  0.4M \\ 
ESPNetv2~\cite{ESPNetv2}  & 0.5 & ESPNetv2    &66.4 & 66.2   & 80     &  0.8M   \\ 
ERFNet~\cite{ERFNet}      & 0.5 &  -          &70.0 &68.0    &  41.9  &  -    \\ 
BiSeNetv1~\cite{bisenet}$^*$   & 0.75 & Xception39  &69.0 &68.4    &  175  &  5.8M    \\

ICNet~\cite{ICnet}      & 1.0   & PSPNet50    &-    &69.5    & 34 &  26.5M  \\
CellNet~\cite{custom_search_seg}&0.75& -      &-     & 70.5 &108 & -      \\
DFANet~\cite{dfanet}     &1.0   & Xception A  &-    &71.3   &100 &7.8M     \\
BiSeNetv2~\cite{bisenetv2}$^*$   & 0.5 & -  &73.4 &72.6   & 28   &  -    \\
DF1-Seg~\cite{DF-seg-net}$^*$&1.0   & DFNet1     & -   &73.0   & 100  & 8.55M   \\
BiSeNetv1~\cite{bisenet}$^*$   & 0.75 & ResNet18    &74.8 &74.7    &  35  &  12.9M    \\ 
DF2-Seg~\cite{DF-seg-net}$^*$& 1.0  &DFNet2      & -   &74.8   & 68  & 18.88M  \\
SwiftNet~\cite{swiftnet}$^*$ & 1.0  &ResNet18     &75.4    &75.8 &39.9&11.8M    \\
FC-HarDNet~\cite{chao2019hardnet}$^*$ & 1.0 & HarDNet & 77.4 &76.0 & 35 & 4.1M \\
SwiftNet-ens~\cite{swiftnet}$^*$&1.0 & -          &-    &76.5   &18.4 &24.7M    \\

\midrule
BiAlignNet    & 0.75 & DFNet2  & 76.8 & 75.4  & 50 & 19.2M \\
BiAlignNet    & 1.0  & DFNet2  & 78.7 & 77.1 & 32 & 19.2M \\
BiAlignNet\textdagger    & 0.75 & DFNet2  & 79.0 & 76.9  & 50 & 19.2M \\
BiAlignNet\textdagger    & 1.0  & DFNet2  & \textbf{80.1} & \textbf{78.5} & 32 & 19.2M \\

\bottomrule[0.1em]
\end{tabular}
}
		\begin{tablenotes}
			 \item {\scriptsize \textdagger Mapillary dataset used for pretraining.
			 }
		\end{tablenotes}
		
	\end{threeparttable}
\end{table}

\begin{figure}[!t]
    \centering
    \includegraphics[width=\linewidth]{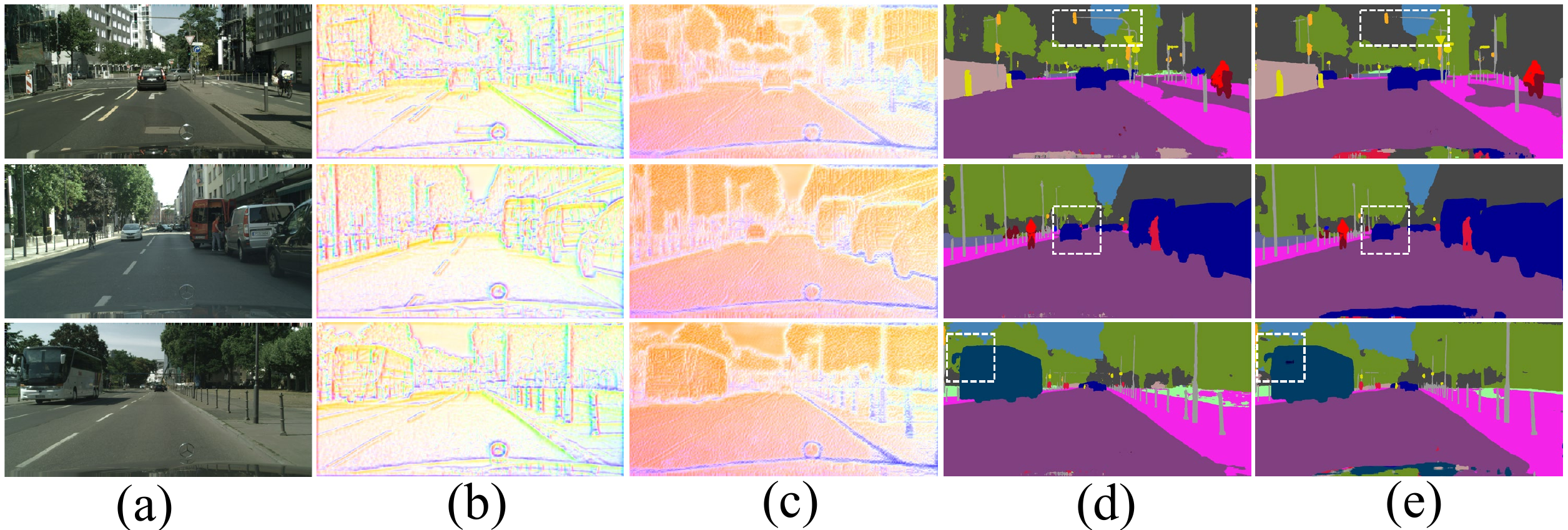}
    \caption{\textbf{Visualization of learned flow field and segmentation output.} Column (a) lists three exemplary images. Column (b) and (c) show the flow field in two directions, spatial to context and context to spatial correspondingly. Column (d) and (e) show the comparison between BiAlignNet and BiSeNet. Best viewed on screen and zoom in.}
    \label{fig:vis}
\end{figure}

\begin{figure}[h]
    \centering
    \includegraphics[width=\linewidth]{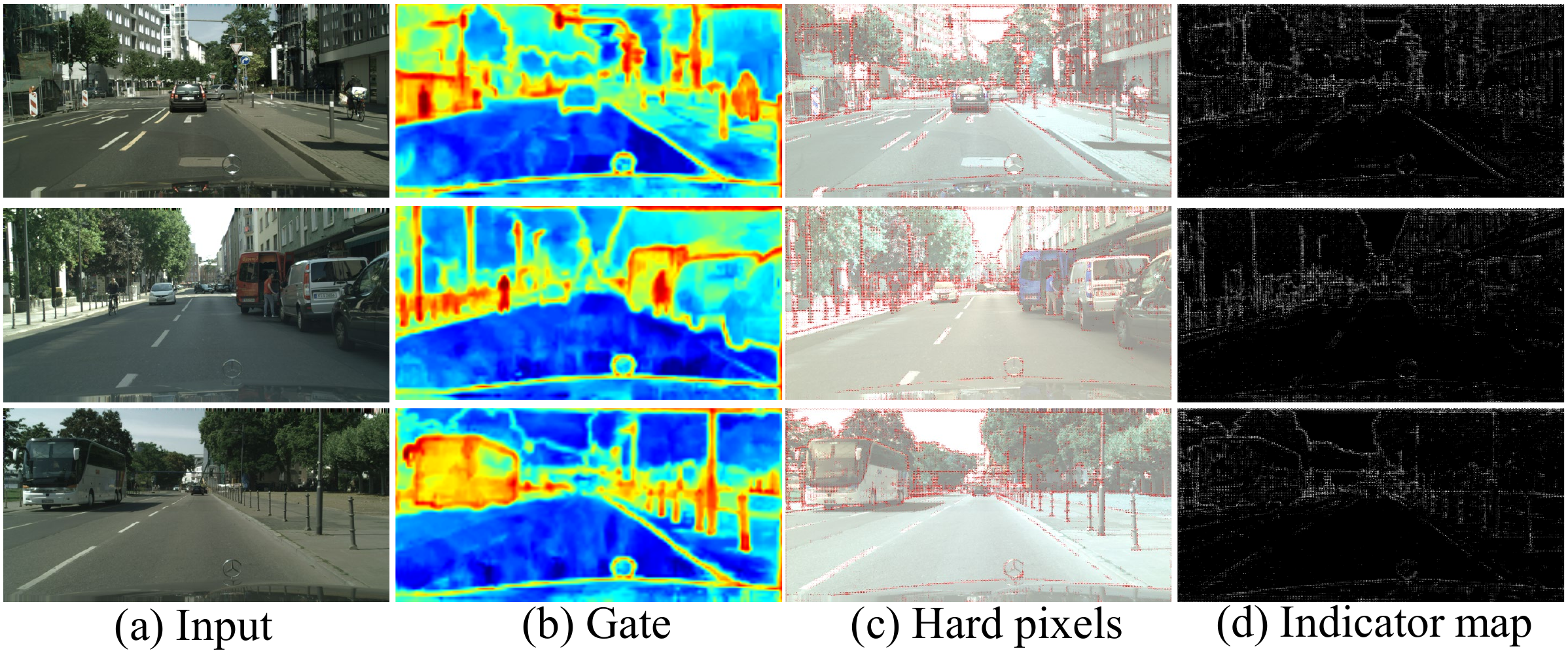}
    \caption{\textbf{Visualization of flow gate, hard examples in spatial loss and predicted edges.} Column (a) lists input images. Column (b) shows the gate map from spatial path to context path. Column (c) shows the hard examples in $L_{hard}$. Column (d) illustrates the predicted hard pixel indicator map from the spatial path. Best viewed on screen and zoom in.}
    \label{fig:gate}
\end{figure}

\noindent
\textbf{Result Comparison.} Table~\ref{table:cityscapes_sota_speed_acc2} shows the results of our method compared to other state-of-the-art real-time methods. 
Our method with an input size of $768\times1536$ can get the best trade-off between accuracy and speed. When input with the whole image, BiAlignNet still runs in real time and gets 78.7\% mIoU and 77.1\% mIoU on val and test, which outperforms all the methods listed above. 
After pre-training on Mapillary~\cite{mapillary} dataset, our BiAlignNet gains 1.4\% improvement. We also apply our method with different light-weight backbones on CamVid dataset and report comparison results in Table~\ref{table:camvid_res}. BiAlignNet also achieves  state-of-the-art performance on the CamVid. \\
\textbf{Visualization.} In Fig.~\ref{fig:vis}, we visualize flow fields from two directions. Flow from the spatial path to the context path (Column b) contains more detailed information and Column c that is from the context path, includes more high-level information. Thus, different features are aligned to each other under the guidance of learned flow field. Fig.~\ref{fig:vis}(d) shows that BiAlignNet outperforms BiSeNet (Column e) on boundaries and details. Fig.~\ref{fig:gate} gives more insights into the proposed GFAM module and the hard pixel mining supervision. As shown in Column b, gates from the spatial path assign higher scores on image details. It confirms that the gate in GFAM can filter the noise and highlight the significant part in the flow field. Fig.~\ref{fig:gate}(c) and (d) visualize hard pixels used in $L_{hard}$ and the predicted indicator map by the spatial path. They are consistent with the fact that edge-guided hard pixel mining pays more attention to fine-grained objects and edges that are difficult to separate. 

\begin{table}[!t]\setlength{\tabcolsep}{10pt}
\caption{\textbf{Comparison on the CamVid {\it test} set with previous state-of-the-art real-time models.}}
		\label{table:camvid_res}
	\centering
	\begin{threeparttable}
		\scalebox{0.70}{
\begin{tabular}{lccc}
\toprule[0.2em]
Method    &  Backbone & mIoU ($\%$)  & \#FPS \\ \toprule[0.2em]
DFANet B~\cite{dfanet} & - & 59.3 & 160 \\
SwiftNet~\cite{swiftnet} & ResNet18 & 63.33 & - \\
DFANet A~\cite{dfanet} & - & 64.7 & 120 \\
 ICNet~\cite{ICnet}&   ResNet-50 & 67.1 & 34.5  \\
 BiSeNetv1~\cite{bisenet} & ResNet18 & 68.7 & 60 \\
 BiSeNetv2~\cite{bisenetv2} & - & 72.4 & 60 \\
 BiSeNetv$\text{2}^*$~\cite{bisenetv2} & - & 76.7 & 60 \\
\midrule 
BiAlignNet & DFNet1 & 68.9  &  85\\
BiAlignNet & DFNet2 &  72.3 &  65\\
BiAlignNe$\text{t}^*$ & DFNet2 &  \textbf{77.1} &  65\\

     \bottomrule[0.1em]
\end{tabular}
}
\begin{tablenotes}
			 \item {\scriptsize * Cityscapes dataset used for pretraining.
			 }
		\end{tablenotes}
	\end{threeparttable}
\end{table}

\subsection{Ablation Study}

We carry out ablation studies on each component of BiAlignNet in this section. As shown in Table~\ref{table:ablation}, our proposed module only introduces a very small amount of computation. 

\noindent
\textbf{Ablation for bidirectional alignment.} We argue that 
insufficiently feature fusion leads to low performance in previous BiSeNet. As we can see in Table~\ref{table:ablation}, compared to the baseline that simply concatenates two feature maps, bidirectional alignment with GFAM can improve performance by 2.4\%. Moreover, the alignments in two directions show the synergistic effects with each other. The performance increase brought by bidirectional alignment is more than the two one-way models.
 Also, the simple gate mechanism in GFAM results in a 0.8\% performance increase. 

\noindent
\textbf{Ablation for the spatial loss.} We expect two paths to learn different contents from the input, especially the spatial path. Thus, we enhance the detail supervision in the spatial path through the specially designed spatial loss with a hard pixel mining indicator. After adding the spatial loss, the performance has improved by 0.9\%. This proves the effectiveness of the designed spatial loss function.  

\begin{table}[!t]\setlength{\tabcolsep}{6pt}
\caption{\textbf{Ablation Study.} We show the  effectiveness of each component in BiAlignNet with DFNet2 on validation set of Cityscapes. \textbf{CP}: Context Path; \textbf{SP}: Spatial Path; \textbf{GFAM}: Gated Flow Alignment Module; \textbf{FAM}: original Flow Alignment Module; $\xrightarrow{}$: Alignment direction; \textbf{SL}: Spatial Loss.} 
\label{table:ablation}
	\centering
	\begin{threeparttable}
		\scalebox{0.65}{
\begin{tabular}{lccc}
\toprule[0.2em]
 Method &  mIoU ($\%$) & $\Delta$ ($\%$) & \#GFLOPs\\ \toprule[0.2em]
CP\,+\,SP (baseline) &  75.4  & -  & 108 \\
CP\,+\,SP\,+\,GFAM (CP$\xrightarrow{}$SP)& 76.5 & 1.1$\uparrow$& 108.37\\
CP\,+\,SP\,+\,GFAM (SP$\xrightarrow{}$CP)& 76.6  &1.2$\uparrow$ & 108.36\\
CP\,+\,SP\,+\,FAM (bidirection)& 77.0  & 1.6$\uparrow$ & 108.72 \\
CP\,+\,SP\,+\,GFAM (bidirection)& 77.8 & 2.4$\uparrow$& 108.73 \\
\midrule
CP\,+\,SP\,+\,GFAM (bidirection)\,+\,SL &  \textbf{78.7} & 3.3$\uparrow$& 108.73\\
     \bottomrule[0.1em]
\end{tabular}
}

\end{threeparttable}
\end{table}

%% file: 5conclusion.tex
\section{Conclusion}
\label{conclusion}
In this paper, we propose a Bidirectional Alignment Network (BiAlignNet) for fast and accurate scene parsing. With the bidirectional alignment and specific supervision in each pathway, the low-level spatial feature can be deeply fused with the high-level context feature. Comparative experiments are performed to show the effectiveness of our proposed components over the baseline models. BiAlignNet also achieves a considerable trade-off between segmentation accuracy and the inference speed. 
